\documentclass[10pt,twocolumn,letterpaper]{article}

\usepackage{iccv}

\definecolor{iccvblue}{rgb}{0.21,0.49,0.74}
\usepackage[pagebackref,breaklinks,colorlinks,allcolors=iccvblue]{hyperref}

\def\paperID{02876} %
\def\confName{ICCV}
\def\confYear{2025}

\title{Benchmarking Burst Super-Resolution for Polarization Images:\\Noise Dataset and Analysis}

\author{Inseung Hwang ~ ~ ~ ~ Kiseok Choi ~ ~ ~ ~ Hyunho Ha ~ ~ ~ ~ Min H. Kim\\
KAIST
}

\newcommand{\mparagraph}[1]{\vspace{0.5mm}\noindent{\textbf{#1.}\hspace{1mm}}}

\usepackage{import} %
\graphicspath{{figs/}}

\usepackage[ruled,noend]{algorithm2e} %

\SetAlFnt{\small}
\SetAlCapFnt{\small}
\SetAlCapNameFnt{\small}
\SetAlCapHSkip{0pt}
\IncMargin{-\parindent}
\usepackage{ifpdf}
\usepackage{graphicx}
\ifpdf
  
\else
\fi
\usepackage{color}
\definecolor{cyan}{cmyk}{1,0,0,0}
\definecolor{darkgreen}{rgb}{0,0.5,0}
\definecolor{orange}{rgb}{1,0.5,0}
\definecolor{magenta}{cmyk}{0,1,0,0}
\definecolor{darkyellow}{cmyk}{0,0,0.75,0}
\definecolor{gray}{rgb}{0.8,0.8,0.8}

\usepackage{amsmath} %

\usepackage[toc,page]{appendix} %
\usepackage{wrapfig} %
\usepackage{comment}
\usepackage{bm}
\usepackage{cuted} %
\usepackage{lipsum} %
\usepackage{mathtools} %
\usepackage{algorithmicx}
\usepackage{algpseudocode}
\usepackage{nicefrac}

\usepackage{gensymb}
\usepackage{float}
\usepackage{soul}
\usepackage{array}
\usepackage{multirow}
\usepackage{hhline}
\usepackage{setspace}
\usepackage{nicefrac}

\algdef{SE}[DOWHILE]{Do}{doWhile}{\algorithmicdo}[1]{\algorithmicwhile\ #1}%

\makeatletter
\renewcommand{\ALG@beginalgorithmic}{\small}
\makeatother

\newcommand{\DELETE}[1]{} %
\newcommand{\IGNORE}[1]{}
\usepackage{datenumber}
\usepackage{calc}
\usepackage[mmddyyyy]{datetime}

\newcounter{datetoday}
\newcounter{diffyears}
\newcounter{diffmonths}
\newcounter{diffdays}

\newcommand{\difftoday}[3]{%
      \setmydatenumber{datetoday}{\the\year}{\the\month}{\the\day}%
      \setmydatenumber{diffdays}{#1}{#2}{#3}%
      \addtocounter{diffdays}{-\thedatetoday}%
      \ifnum\value{diffdays}>0
        \def\diffbefore{}%
        \def\diffafter{left}%
      \else
        \def\diffbefore{}%
        \def\diffafter{ago}%
        \setcounter{diffdays}{-\value{diffdays}}%
      \fi
      \setcounter{diffyears}{\value{diffdays}/365}%
      \setcounter{diffdays}{\value{diffdays}-365*\value{diffyears}}%
      \setcounter{diffmonths}{\value{diffdays}/30}%
      \setcounter{diffdays}{\value{diffdays}-30*\value{diffmonths}}%
      \diffbefore
      \ifnum\value{diffyears}=0
      \else
        \ifnum\value{diffyears}>1
            \thediffyears\space years,
        \else
            \thediffyears\space year,
        \fi
      \fi
      \ifnum\value{diffmonths}=0
      \else
        \ifnum\value{diffmonths}>1
            \thediffmonths\space months
        \else
            \thediffmonths\space month
        \fi
      \fi
      \ifnum\value{diffdays}=0
      \else
        \ifnum\value{diffdays}>1
            \thediffdays\space days
        \else
            \thediffdays\space day
        \fi
      \fi
      \diffafter
}

\begin{document}
\maketitle
\begin{abstract}
\noindent
Snapshot polarization imaging calculates polarization states from linearly polarized subimages. To achieve this, a polarization camera employs a double Bayer-patterned sensor to capture both color and polarization. It demonstrates low light efficiency and low spatial resolution, resulting in increased noise and compromised polarization measurements. Although burst super-resolution effectively reduces noise and enhances spatial resolution, applying it to polarization imaging poses challenges due to the lack of tailored datasets and reliable ground truth noise statistics. To address these issues, we introduce PolarNS and PolarBurstSR, two innovative datasets developed specifically for polarization imaging. PolarNS provides characterization of polarization noise statistics, facilitating thorough analysis, while PolarBurstSR functions as a benchmark for burst super-resolution in polarization images. These datasets, collected under various real-world conditions, enable comprehensive evaluation. Additionally, we present a model for analyzing polarization noise to quantify noise propagation, tested on a large dataset captured in a darkroom environment. As part of our application, we compare the latest burst super-resolution models, highlighting the advantages of training tailored to polarization compared to RGB-based methods. This work establishes a benchmark for polarization burst super-resolution and offers critical insights into noise propagation, thereby enhancing polarization image reconstruction.

\end{abstract}
    
\section{Introduction}
\label{sec:intro}
\begin{figure}[t]
	\vspace{-4mm}
	\centering
	\includegraphics[width=\linewidth]{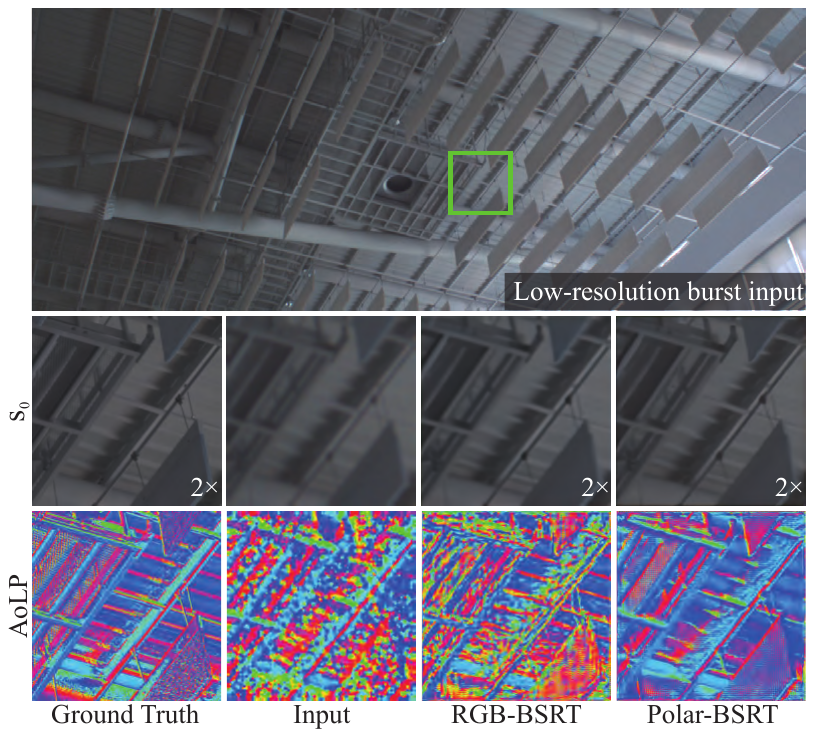}%
	\vspace{-4mm}
	\caption{\label{fig:teaser}%
	The top row displays the low-resolution burst input captured using a polarization camera. The second row presents close-up comparisons of different methods: the ground truth, the low-resolution input, the 2$\times$ super-resolution results of the RGB-trained BSRT method~\cite{luo2022bsrt} trained on a conventional RGB dataset, and the Polar-BSRT method trained on our polarization image dataset. Training with our dedicated polarization dataset significantly improves both the spatial resolution of the intensity image ($s_0$) and the angle of linear polarization (AoLP) map, demonstrating the importance of polarization-specific training for burst super-resolution.}
	\vspace{-5mm}
\end{figure}

Polarization imaging provides valuable scene information by capturing the polarization state of light, which changes upon interaction with surfaces according to the Fresnel equations \cite{Atkinson2006}. 
This property enables polarization imaging to trace light sources and infer surface characteristics, making it widely applicable in areas, such as 
dehazing~\cite{Schechner01, Fang:14, Liu:15, Zhou21}, transparent object segmentation~\cite{Kalra20, Mei2022, Qiao_2023_ICCV}, enhancement of ToF imaging~\cite{Baek:PolarLT:2021, Baek:PolarToF:2022, Jeon_2023_CVPR, scheuble2024pollidar}, reflection removal~\cite{Kong14, eccv2018/Wieschollek, lyu2019_polarRS, Lei_2020_CVPR}, specular-diffuse separation~\cite{Ma07, Ghosh10, Riviere17}, facial acquisition~\cite{Ghosh11, Gotardo18, Riviere20, azinovic2022polface, polarface:SIGA:2024},
shape-from-polarization techniques~\cite{Miyazaki03, Atkinson2006, Huynh13}, 
RGB-D integration~\cite{kadambi2015polar}, normal regularization~\cite{Smith16, Tozza2017}, multiview acquisition~\cite{cui2017polarimetric, Cui2019, Zhu2019, Zhao20}, deep learning approaches~\cite{Ba20, Deschaintre21, Ding2021, Fukao2021, Lei22, Dave22}, and neural implicit representations~\cite{Yufei2024, Li_NeISF_CVPR2024}.
However, polarization-based measurements are highly susceptible to noise, which significantly affects their accuracy and reliability. 
Unlike standard intensity images, polarization properties are not captured directly but are instead derived from multiple intensity measurements through computational models. 
This process inherently amplifies noise, making polarization-sensitive analysis particularly challenging.

The propagation of noise in polarization calculations is complex, often requiring approximations such as first-order Taylor expansions \cite{Goudail:10,Dai:18,Li:19,Roussel:18,Chen:21,Bai:22} or assumptions that noise follows an additive white Gaussian distribution in the Stokes vector components \cite{simmons1985point,quinn2012bayesian,plaszczynski14,kupinski14}.
The lack of precise noise characterization limits the development of robust denoising methods. 
Moreover, modern snapshot polarization cameras employ a double Bayer-patterned sensor to capture both color and polarization information, resulting in lower light efficiency and reduced spatial resolution. These factors collectively lead to increased noise, further degrading polarization-based analysis.

While conventional burst super-resolution (BSR) techniques have successfully mitigated noise and enhanced resolution in standard intensity images, their application to polarization imaging remains unexplored because there are no dedicated training datasets available. 
Without reliable polarization burst super-resolution datasets or detailed noise statistics, it is impossible to train or evaluate polarization-aware super-resolution models effectively.

To address these challenges, we introduce two novel datasets specifically designed for polarization imaging: PolarNS and PolarBurstSR. PolarNS is the first large-scale dataset focused on polarization noise statistics, providing a detailed characterization of noise propagation in polarization images. Captured under diverse real-world conditions with high precision, it serves as a benchmark for analyzing polarization noise and evaluating denoising techniques. 
To complement this dataset, we propose a polarization noise analysis model that quantifies noise propagation through an analytical approach with minimal assumptions, based on a shot and read noise model of polarization sensors. 
This model estimates the statistical distribution of polarization properties in the presence of noise, offering a physically grounded validation of the dataset.

In addition, we establish PolarBurstSR, a dataset designed for training and evaluating burst super-resolution (BSR) models in polarization imaging. By providing high-quality image sequences tailored to polarization data, PolarBurstSR enables the development of BSR techniques that effectively enhance spatial resolution and reduce noise. 
See Figure~\ref{fig:teaser} for an example.
To promote further research, we will publicly release our datasets, along with pretrained models and training pipelines for burst super-resolution on polarization images.

\section{Related Work}
\label{sec:related_work}

\mparagraph{Noise Analysis for Polarization Imaging}  
Traditional noise analysis in polarization imaging has focused on controlled ellipsometric setups with strong polarization signals, multiple image captures, and polarized lighting~\cite{Goldstein:90, Sabatke:00, Takakura:07, Gamiz99, Perkins:10, Goudail:09}.  
Recent work has extended to noise propagation analysis in polarization properties such as the degree of linear polarization (DoLP) and angle of linear polarization (AoLP), often using first-order Taylor expansions~\cite{Goudail:10, Dai:18, Li:19, Roussel:18, Chen:21, Bai:22}.  
However, these methods assume high signal-to-noise ratios and do not generalize well to real-world vision tasks where polarization signals are weak and noise dominates.  
Unlike previous approaches, our work introduces a snapshot polarization imaging noise model tailored for four-channel linear polarization cameras in uncontrolled settings.  
By directly modeling noise propagation with minimal assumptions, our method provides a more comprehensive characterization, capturing not just variance but the full distribution and bias of polarization properties.

\mparagraph{Polarization Image Datasets}  
Polarization datasets have been developed for high-level vision tasks such as segmentation, geometry estimation, and reflection removal~\cite{Liang22, Mei2022, Ba20, Dave22, Lei22, Tian23, Lei_2020_CVPR, Jeon_2024_CVPR}, typically using off-the-shelf polarization cameras.  
Low-level tasks, such as demosaicing and sparse filter design, require full-frame polarization data captured using rotating polarizers~\cite{Qiu21, Kurita_2023_WACV}, though these setups are impractical for hand-held capture.  
In contrast, we introduce a new polarization image dataset specifically designed for burst super-resolution and noise analysis.  
Our dataset provides both image variance and metadata, making it uniquely suited for denoising and super-resolution tasks.  
To ensure high-quality ground truth, we employ a burst shot method, offering 14 low-resolution burst images per scene alongside a high-resolution and noise suppressed reference.  
This dataset serves as a benchmark for polarization burst super-resolution, addressing a critical gap in existing resources.

\mparagraph{Polarization Imaging Applications}  
Polarization imaging is widely used in applications such as dehazing~\cite{Schechner01, Fang:14, Liu:15, Zhou21}, reflection removal~\cite{Kong14, eccv2018/Wieschollek, lyu2019_polarRS, Lei_2020_CVPR}, transparent object segmentation~\cite{Kalra20, Mei2022, Qiao_2023_ICCV}, and ToF imaging enhancement~\cite{Baek:PolarLT:2021, Baek:PolarToF:2022, Jeon_2023_CVPR, scheuble2024pollidar}.  
While specular reflection provides strong polarization signals, diffuse polarization is weak and directly affected by noise, making applications like shape-from-polarization~\cite{Miyazaki03, Atkinson2006, Huynh13} highly sensitive to noise.  
Our work introduces a dedicated noise model for DoLP and AoLP, addressing the lack of systematic noise characterization in diffuse polarization imaging.  
Additionally, our dataset includes real-world scenes such as glass-through, reflection-dominant, and road surface imaging, where burst super-resolution significantly reduces noise.

\mparagraph{Multi-Frame Super-Resolution}  
Multi-frame super-resolution (SR) has evolved from early algorithmic approaches~\cite{Tsai84, irani91, elad97, Farsiu04} to modern burst super-resolution (BSR) methods that leverage hand-held motion~\cite{Wronski19}.  
Learning-based BSR was first introduced by Bhat et al.~\cite{Bhat2021}, whose dataset and evaluation metrics have been widely adopted~\cite{Bhat_2021_ICCV, Dudhane_2023_CVPR, luo2022bsrt, Wei_2023_ICCV, Kang24, Bhat23}.  
Recent expansions include self-supervised learning and datasets with improved scene variety~\cite{Bhat23, Wei_2023_ICCV}.  
However, burst super-resolution has not been explored for polarization imaging, where noise characteristics differ significantly from RGB images.  
We introduce PolarBurstSR, the first burst super-resolution dataset and benchmarking framework for polarization imaging, enabling denoising and high-resolution reconstruction in this domain.

\section{Dataset}
\label{sec:dataset}

\subsection{Polarization Noise Statistic Dataset}  
\begin{figure}
\centering
\def\svgscale{1}
\footnotesize
\graphicspath{{figs/}}
\begingroup%
  \makeatletter%
  \providecommand\color[2][]{%
    \errmessage{(Inkscape) Color is used for the text in Inkscape, but the package 'color.sty' is not loaded}%
    \renewcommand\color[2][]{}%
  }%
  \providecommand\transparent[1]{%
    \errmessage{(Inkscape) Transparency is used (non-zero) for the text in Inkscape, but the package 'transparent.sty' is not loaded}%
    \renewcommand\transparent[1]{}%
  }%
  \providecommand\rotatebox[2]{#2}%
  \newcommand*\fsize{\dimexpr\f@size pt\relax}%
  \newcommand*\lineheight[1]{\fontsize{\fsize}{#1\fsize}\selectfont}%
  \ifx\svgwidth\undefined%
    \setlength{\unitlength}{229.77167805bp}%
    \ifx\svgscale\undefined%
      \relax%
    \else%
      \setlength{\unitlength}{\unitlength * \real{\svgscale}}%
    \fi%
  \else%
    \setlength{\unitlength}{\svgwidth}%
  \fi%
  \global\let\svgwidth\undefined%
  \global\let\svgscale\undefined%
  \makeatother%
  \begin{picture}(1,0.53395879)%
    \lineheight{1}%
    \setlength\tabcolsep{0pt}%
    \put(0,0){\includegraphics[width=\unitlength,page=1]{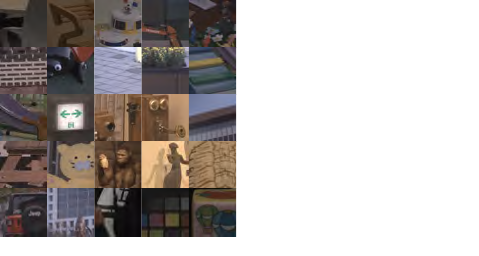}}%
    \put(0.4982028,0.00539346){\makebox(0,0)[t]{\lineheight{1.25}\smash{\begin{tabular}[t]{c}
    			\end{tabular}}}}%
    \put(0,0){\includegraphics[width=\unitlength,page=2]{dataset.pdf}}%
  \end{picture}%
\endgroup%

\vspace{-2em}
\caption{\label{fig:dataset_thumbnail}
Our dataset examples presented in color intensity images $s_{0}$ (left) and their DoLP properties (right).}
\vspace{-3mm}
\end{figure}

We introduce the noise-reduced image and the noise statistic dataset, called PolarNS. We capture this data using an off-the-shelf polarization RGB camera (FLIR BFS-U3-51S5PC-C) and a 35mm lens. The noise statistics of each scene are calculated from the burst images taken in a fixed position, ensuring that all our scenes remain static. Our scenes consist of 54 objects in a darkroom and 190 indoor and outdoor scenes. We capture 10,000 images for each object in the darkroom and around 1,000 images for each indoor and outdoor scene. The illumination also remains static throughout the capture period. The exposure time is set as a multiple of the AC current period for the indoor scenes. For the outdoor scenes, occlusion caused by clouds leads to variations in capture time, so we capture the scenes when there are no clouds in the sky or at nighttime. We calculate the mean and variance per pixel from the raw burst images. The dataset includes the mean and variance of the scenes, along with metadata (e.g., exposure time, gain) for the images. Figure~\ref{fig:dataset_thumbnail} shows examples of our datasets.

\subsection{Burst Super-Resolution Dataset}  
We also acquire a polarization burst super-resolution dataset called PolarBurstSR. We follow the same process as the BurstSR dataset~\cite{Bhat2021} using a polarization camera. The ground truth of the burst dataset is derived from the mean image of the noise statistic dataset. When the noise statistics are captured, 14 hand-held burst images are taken simultaneously. 
The total number of images is 160, distributed among the training, validation, and test sets as 112, 16, and 32 respectively.  
Refer to the supplemental document for the super-resolution image processing details.

\section{Noise analysis model}
\label{sec:noise_model}
We derive a new noise model for the Stokes vector and its polarization properties, specifically designed for a conventional image denoising and super-resolution framework. Key physical findings from this noise model include:
\begin{itemize}
	\item The variance of the Stokes vector components depends linearly on the $s_0$ component of the Stokes vector.
	\item The degree of linear polarization follows a Rician distribution influenced by the intensity of the $s_0$ component, polarization intensity, and Stokes vector noise.  
	\item The angle of linear polarization exhibits a Gaussian-like distribution shaped by the ratio of polarization intensity to Stokes vector noise.
\end{itemize}
In the following subsections, we will demonstrate each of these properties in detail.

\subsection{Background}
The light is an electromagnetic wave in which the electric field oscillates perpendicularly to the direction of wave propagation. This perpendicular oscillation allows for a degree of freedom in the oscillation, which gives light its orientation called polarization. The electric field can oscillate linearly, circularly, or elliptically, depending on the orientation and the retardation of the perpendicular components. The Stokes vector 
$\mathbf{s}=\left[ \begin{matrix}
	{{s}_{0}} & {{s}_{1}} & {{s}_{2}} & {{s}_{3}} 
\end{matrix} \right]_{{}}^{T}$
is the four-dimensional vector, which represents the polarization state of light. $s_0$ indicates the total intensity of light, while $s_1$, $s_2$, and $s_3$ refer to the intensities of the polarized components, describing horizontal/vertical, diagonal/antidiagonal, and circular polarization, respectively. In this paper, we focus solely on linearly polarized light because the employed polarization camera can capture linear polarization only; therefore, we do not utilize $s_3$ and represent the Stokes vector as a three-dimensional vector.

From polarization, we can obtain useful properties: the degree of linear polarization and the angle of linear polarization. These properties provide intuitive information about polarization. DoLP is the ratio of the intensity of linear polarization to the total intensity. DoLP, denoted as $\psi$, is defined as:
	$\psi =\frac{{{s}_{pol}}}{{{s}_{0}}}=\frac{\sqrt{s_{1}^{2}+s_{2}^{2}}}{{{s}_{0}}}$,
where $s_{pol}$ represents the intensity of the polarized components, indicating the degree of light polarization.

The AoLP refers to the orientation of linear polarization. The AoLP, denoted as $\phi$, is defined as:
	$\phi =\frac{1}{2}{{\tan }^{-1}}(\frac{s_{2}^{{}}}{{{s}_{1}}})$.
It indicates the orientation of the surface normal during diffuse reflection.

To acquire the linear components of the Stokes vector, at least three different angle linear polarization images are required. An off-the-shelf polarization camera uses four angles of linear polarization, specifically 0, 45, 90, and 135 degrees, and we will derive the equations in the case of these four angles in this paper. 
The polarization images $I_{\left(\cdot\right)}$ can be expressed as
\begin{equation}
\label{eq:polarization_acquisition}
	\resizebox{0.7\linewidth}{!}{
	\mbox{\fontsize{10}{12}\selectfont $
	\left[ \begin{matrix}
		{{I}_{0}}  \\
		{{I}_{45}}  \\
		{{I}_{90}}  \\
		{{I}_{135}}  \\
	\end{matrix} \right]=\frac{1}{2}\left[ \begin{matrix}
		1 & 1 & 0  \\
		1 & 0 & 1  \\
		1 & -1 & 0  \\
		1 & 0 & -1  \\
	\end{matrix} \right]\left[ \begin{matrix}
		{{s}_{0}}  \\
		{{s}_{1}}  \\
		{{s}_{2}}  \\
	\end{matrix} \right].
$ } } %
\end{equation}
The reconstructed Stokes vector $\mathbf{\hat{s}}$ is expressed as
\begin{equation}
\label{eq:stokes_vector_recon}
	\resizebox{0.65\linewidth}{!}{
	\mbox{\fontsize{10}{12}\selectfont $
	\mathbf{\hat{s}}=\left[ \begin{matrix}
		\frac{1}{2}\left( {{{\hat{I}}}_{0}}+{{{\hat{I}}}_{45}}+{{{\hat{I}}}_{90}}+{{{\hat{I}}}_{135}} \right)  \\
		{{{\hat{I}}}_{0}}-{{{\hat{I}}}_{90}}  \\
		{{{\hat{I}}}_{45}}-{{{\hat{I}}}_{135}}  \\
	\end{matrix} \right],
$ } } %
\end{equation}
from observed images $\hat{I}_{\left(\cdot\right)}$. 

\subsection{Stokes vector noise model}
Image noise has two major categories: shot noise and read noise. Shot noise arises from the quantum properties of light. It follows a Poisson distribution regarding the number of photons and can be approximated by a Gaussian distribution. Read noise consists of residual noise sources, such as thermal noise, source follower noise, and ADC quantization noise. In many applications, read noise is modeled as zero-mean Gaussian noise. The observed image $\hat{I}$ can be modeled as the random variable:
\begin{equation}
	\label{eq:image_noise_model}
	\hat{I}\sim\mathcal{N}\left( I,I\sigma _{s}^{2}+\sigma _{r}^{2} \right),
\end{equation}
where $\sigma_{s}$ is the coefficient of shot noise, and $\sigma_{r}$ is the standard deviation of read noise.

Equations~\eqref{eq:polarization_acquisition} and \eqref{eq:image_noise_model} express the linear polarization image noise model as 
\begin{align}
	\label{eq:polar_image_noise_model}
	{{{\hat{I}}}_{0}}&\sim\mathcal{N}\left( \frac{1}{2}{{s}_{0}}+\frac{1}{2}{{s}_{1}},\left( \frac{1}{2}{{s}_{0}}+\frac{1}{2}{{s}_{1}} \right)\sigma _{s}^{2}+\sigma _{r}^{2} \right), \nonumber \\ 
	{{{\hat{I}}}_{45}}&\sim\mathcal{N}\left( \frac{1}{2}{{s}_{0}}+\frac{1}{2}{{s}_{2}},\left( \frac{1}{2}{{s}_{0}}+\frac{1}{2}{{s}_{2}} \right)\sigma _{s}^{2}+\sigma _{r}^{2} \right), \nonumber \\ 
	{{{\hat{I}}}_{90}}&\sim\mathcal{N}\left( \frac{1}{2}{{s}_{0}}-\frac{1}{2}{{s}_{1}},\left( \frac{1}{2}{{s}_{0}}-\frac{1}{2}{{s}_{1}} \right)\sigma _{s}^{2}+\sigma _{r}^{2} \right), \text{and} \nonumber \\ 
	{{{\hat{I}}}_{135}}&\sim\mathcal{N}\left( \frac{1}{2}{{s}_{0}}-\frac{1}{2}{{s}_{2}},\left( \frac{1}{2}{{s}_{0}}-\frac{1}{2}{{s}_{2}} \right)\sigma _{s}^{2}+\sigma _{r}^{2} \right).
\end{align}
Each image is an independent random variable. The random variables are independent and normally distributed, so they can be subjected to arithmetic operations. By Equations~\eqref{eq:stokes_vector_recon} and \eqref{eq:polar_image_noise_model}, the Stokes vector model is expressed as
\begin{equation}
\label{eq:stokes_noise_model}
	\resizebox{0.95\linewidth}{!}{
	\mbox{\fontsize{10}{12}\selectfont $
	{{{\hat{s}}}_{0}} \sim\mathcal{N}\left( {{s}_{0}},\frac{1}{2}\sigma _{v}^{2} \right), ~ ~
	{{{\hat{s}}}_{1}} \sim\mathcal{N}\left( {{s}_{1}},\sigma _{v}^{2} \right),  ~ ~
	{{{\hat{s}}}_{2}} \sim\mathcal{N}\left( {{s}_{2}},\sigma _{v}^{2} \right),
$ } } %
\end{equation}
where Stokes vector noise $\sigma _{v}^{2}={{s}_{0}}\sigma _{s}^{2}+2\sigma _{r}^{2}$. 

Note that ${\hat{s}}_0$ depends on ${\hat{s}}_1$ and ${\hat{s}}_2$, while ${\hat{s}}_1$ and ${\hat{s}}_2$ are independent of one another. The following sections are based on this independence, which exists only in the case of the four angles.

\subsection{DoLP noise model}

DoLP is the ratio of the polarization components to the total intensity. The distributions of DoLP and AoLP, when the polarization component follows a Gaussian distribution, are known from previous astronomical studies \cite{simmons1985point,quinn2012bayesian}. Intuitively, the distribution of the DoLP represents the marginal distribution of the radius of the noncentral Gaussian distribution, which corresponds to the line integral along the blue circle in Figure~\ref{fig:polar_coordinate}(a).

The division of the random variables makes it difficult to derive the equations. Because the noise of total intensity is much smaller than the total intensity, we approximate the observed total intensity ${\hat{s}}_0$ to the true total intensity $s_0$. Then, the normalized Stokes vector components are
\begin{equation}
\label{eq:stokes_approx}
	\resizebox{0.95\linewidth}{!}{
	\mbox{\fontsize{10}{12}\selectfont $
	{{{\hat{s}}}_{1}}/{{s}_{0}} \sim\mathcal{N}\left( {{s}_{1}}/{{s}_{0}},\sigma _{v}^{2}/s_{0}^{2} \right), ~  ~
	{{{\hat{s}}}_{2}}/{{s}_{0}} \sim\mathcal{N}\left( {{s}_{2}}/{{s}_{0}},\sigma _{v}^{2}/s_{0}^{2} \right).
$ } } %
\end{equation}

The components of the normalized Stokes vector follow a Gaussian distribution. 
From the DoLP equation and Equation~\eqref{eq:stokes_approx}, 
the DoLP is distributed as a Rician distribution. It is expressed as
\begin{equation}
	\label{eq:dolp_model}
	\hat{\psi }\sim\text{Rice}\left( \psi ,{{\sigma }_{v}}/{{s}_{0}} \right).
\end{equation}

The probability density function (PDF) of DoLP is expressed as
\begin{align}
	\label{eq:dolp_pdf}
	&{{f}_{\text{Rice}}}\left( \hat{\psi }|\psi ,{{\sigma }_{v}}/{{s}_{0}} \right) \nonumber \\
	&=\frac{s_{0}^{2}\hat{\psi }}{\sigma _{v}^{2}}\exp \left( \frac{-s_{0}^{2}\left( \hat{\psi }_{{}}^{2}+\psi _{{}}^{2} \right)}{2\sigma _{v}^{2}} \right){{I}_{0}}\left( \frac{s_{0}^{2}\hat{\psi }\psi }{\sigma _{v}^{2}} \right).
\end{align}
The mean of the distribution is represented as
\begin{equation}
	\label{eq:dolp_mean}
	\text{E}\left[ {\hat{\psi }} \right]=\frac{\sigma _{v}^{{}}}{s_{0}^{{}}}\sqrt{\frac{\pi }{2}}{{L}_{1/2}}\left( \frac{-s_{0}^{2}\psi _{{}}^{2}}{2\sigma _{v}^{2}} \right).
\end{equation}

Figures~\ref{fig:polar_coordinate}(b) and (c) exemplify the distribution of DoLP and its associated bias from the parameters, specifically the DoLP and the ratio of $s_0$ to Stokes vector noise.

\subsection{AoLP noise model}
\begin{figure}[t]
	\centering
	\includegraphics[width=\linewidth]{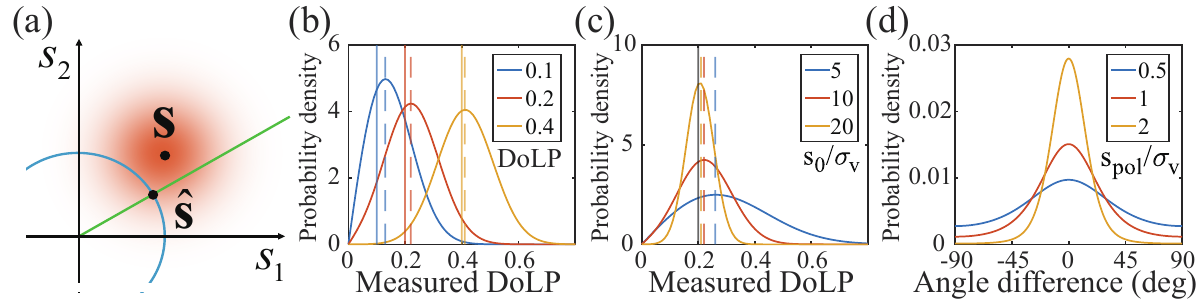}%
	\vspace{-3mm}%
	\caption{\label{fig:polar_coordinate}%
	(a) Description of the noise distribution: The noise distribution of a real vector $\mathbf{s}$ is illustrated in red, while the DoLP distribution for an arbitrary observation $\hat{\mathbf{s}}$ is depicted as the line integral of the blue circle. The AoLP distribution corresponds to the integral of the green line. (b) Distribution of observed DoLP in relation to true DoLP: The vertical solid line signifies the true DoLP, and the vertical dashed line indicates the biased maximum. (c) Distribution of observed DoLP concerning the ratio of $s_0$ and Stokes vector noise: The vertical solid line represents the true DoLP, while the vertical dashed line symbolizes the biased maximum. (d) Distribution of observed AoLP relative to the ratio of $s_{pol}$ and the Stokes vector noise.}
	\vspace{-3mm}%
\end{figure}

The noise model of AoLP represents the marginal distribution of the phase of ${\hat{s}}_1$ and ${\hat{s}}_2$. In Figure~\ref{fig:polar_coordinate}(a), we observe the line integral along the green line. The PDF of AoLP is expressed as
\begin{align}
	\label{eq:aolp_pdf}
	&f_{\phi }^{{}}\left( 2\hat{\phi }|{{s}_{1}},{{s}_{2}},\sigma _{v}^{2} \right)=\int_{0}^{\infty }{f\left( {{{\hat{s}}}_{1}},{{{\hat{s}}}_{2}} \right)rdr} \nonumber \\ 
	& =\frac{1}{2\pi }\exp \left( -\frac{s_{pol}^{2}}{2\sigma _{v}^{2}} \right)+\frac{s_{pol}^{{}}\cos \left( 2\hat{\phi }-2\phi  \right)}{\sqrt{2\pi }\sigma _{v}^{{}}} \nonumber \\
	&\cdot \exp \left( -\frac{s_{pol}^{2}{{\sin }^{2}}\left( 2\hat{\phi }-2\phi  \right)}{2\sigma _{v}^{2}} \right)\Phi \left( \frac{s_{pol}^{{}}\cos \left( 2\hat{\phi }-2\phi  \right)}{\sigma _{v}} \right),
\end{align}
where $\Phi$ is the cumulative distribution function of the standard normal distribution. The entire integration process is detailed in the supplemental material.
The PDF is symmetric about the true AoLP, making the mean unbiased. If the ratio of the polarization intensity to the Stokes vector noise is zero, the distribution becomes a uniform distribution with a density $1/\pi$. If the polarization intensity is much larger than the Stokes vector noise ($s_{pol}\gg\sigma _{v}$), it can be approximated by a Gaussian distribution. Figure~\ref{fig:polar_coordinate}(d) illustrates an example of the AoLP distribution based on the ratio of $s_{pol}$ to Stokes vector noise.

\section{Noise analysis}
\label{sec:validation}

\subsection{Validation for noise analysis model}
We validate the noise analysis model using object scenes in a darkroom. We obtain pseudo true values from the mean of 10,000 burst images. The observed values are the pixel values of each image from the scenes. Our noise distribution model consists of three parts, and we display the distribution of these parts using a histogram.
For validation of the Stokes vector noise model, refer to the supplemental material.

\begin{figure*}[t]
	\centering
	\includegraphics[width=\linewidth]{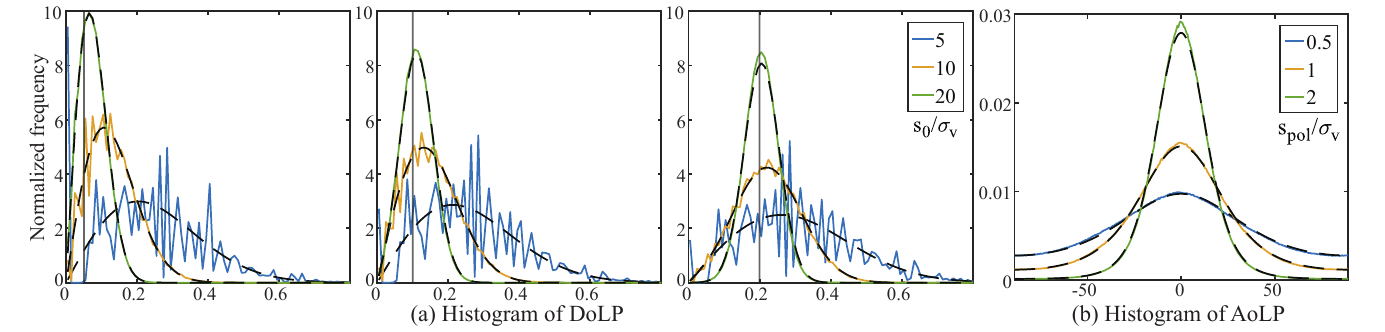}%
	\vspace{-1mm}%
	\caption{\label{fig:dolp_histogram}
	Histogram of polarization properties. The histogram values are normalized to a probability density scale. The black dashed line indicates the distribution predicted by our model. Each colored line represents the measured DoLP distribution at the corresponding ratio of the Stokes component value to the Stokes vector noise. (a) Histogram of DoLP. Each histogram displays the distribution for true DoLP values of 0.05, 0.1, and 0.2, respectively. The vertical line marks the location of the true DoLP value. (b) Histogram of AoLP.}
	\vspace{-3mm}%
\end{figure*}

We obtain the observed distribution of DoLP and AoLP differences using a histogram. The DoLP distribution is a function of the true DoLP and the signal-to-noise ratio of $s_0$. To validate the model for each parameter, a 3D histogram is required, based on the true DoLP, the SNR of $s_0$, and the observed DoLP value. The differences between the true AoLP and the observed AoLP are distributed as a function of the SNR of $s_{pol}$. Similarly to DoLP, we obtain a 2D histogram from the samples. 

The results are illustrated in Figure \ref{fig:dolp_histogram}. The colored lines represent the observed histogram for each SNR, while the black lines depict our estimated probability density functions. The histograms are normalized by the sample size and the bin width to match the scale of the probability density function. The estimated distribution is not merely the outcome of curve fitting to the observed histogram; it reflects data that closely resembles the observed values. For the same true DoLP, the biases vary with SNR, as estimated in Equation~\eqref{eq:dolp_mean}.

\subsection{Comparison with polarization datasets}
We compare the noise levels of our GT dataset with those of other polarization image datasets. Our dataset is the first to provide noise statistics for polarization image datasets; thus, prior datasets do not calculate or provide their own noise statistics. We cannot make direct comparisons using noise statistics; therefore, we estimate the noise level from the $s_0$ images utilizing a single-image noise level estimation method for RGB images~\cite{Chen2015}.

We select general scene polarization RGB image datasets for comparison: the KAUST polarization image dataset~\cite{Qiu21}, the RSP dataset~\cite{Kurita_2023_WACV}, the MCubeS dataset~\cite{Liang22}, and the spectro-polarimetric dataset~\cite{Jeon_2024_CVPR}.
We focus exclusively on real captured, visible spectrum, and RGB images, thus excluding synthetic, near-infrared, or hyperspectral images from these datasets. 
The results of the noise level estimation are presented in Table \ref{tab:single_image_noise_level}. Our dataset demonstrates the lowest estimated noise level in both the absolute mean and squared mean, as indicated in PSNR.

\begin{table}
	\centering
	\caption{\label{tab:single_image_noise_level}
	Comparison among the polarized image datasets through noise level estimation. Bold text signifies the result with the least noise in that metric.}
	\vspace{-2mm}
	\resizebox{0.98\linewidth}{!}{
		\begin{tabular}{lcc}
			\toprule
			Dataset & Mean (std.dev.)$\times10^{3}\downarrow$ & PSNR$\uparrow$ \\
			\midrule
			KAUST~\cite{Qiu21} & 2.733 (1.383) & 50.30 \\
			RSP~\cite{Kurita_2023_WACV} & 1.747 (2.231) & 50.96 \\
			MCubeS~\cite{Liang22} & 0.973 (0.369) & 59.65 \\
			Spectro-polarimetric~\cite{Jeon_2024_CVPR} & 0.346 (0.252) & 67.37 \\
			Our dataset & \textbf{0.056 (0.076)} & \textbf{80.50} \\
			\bottomrule
	\end{tabular}}
	\vspace{-2mm}	
\end{table}

\subsection{Statistics of our dataset}
\begin{figure}[t]
	\centering
	\includegraphics[width=\linewidth]{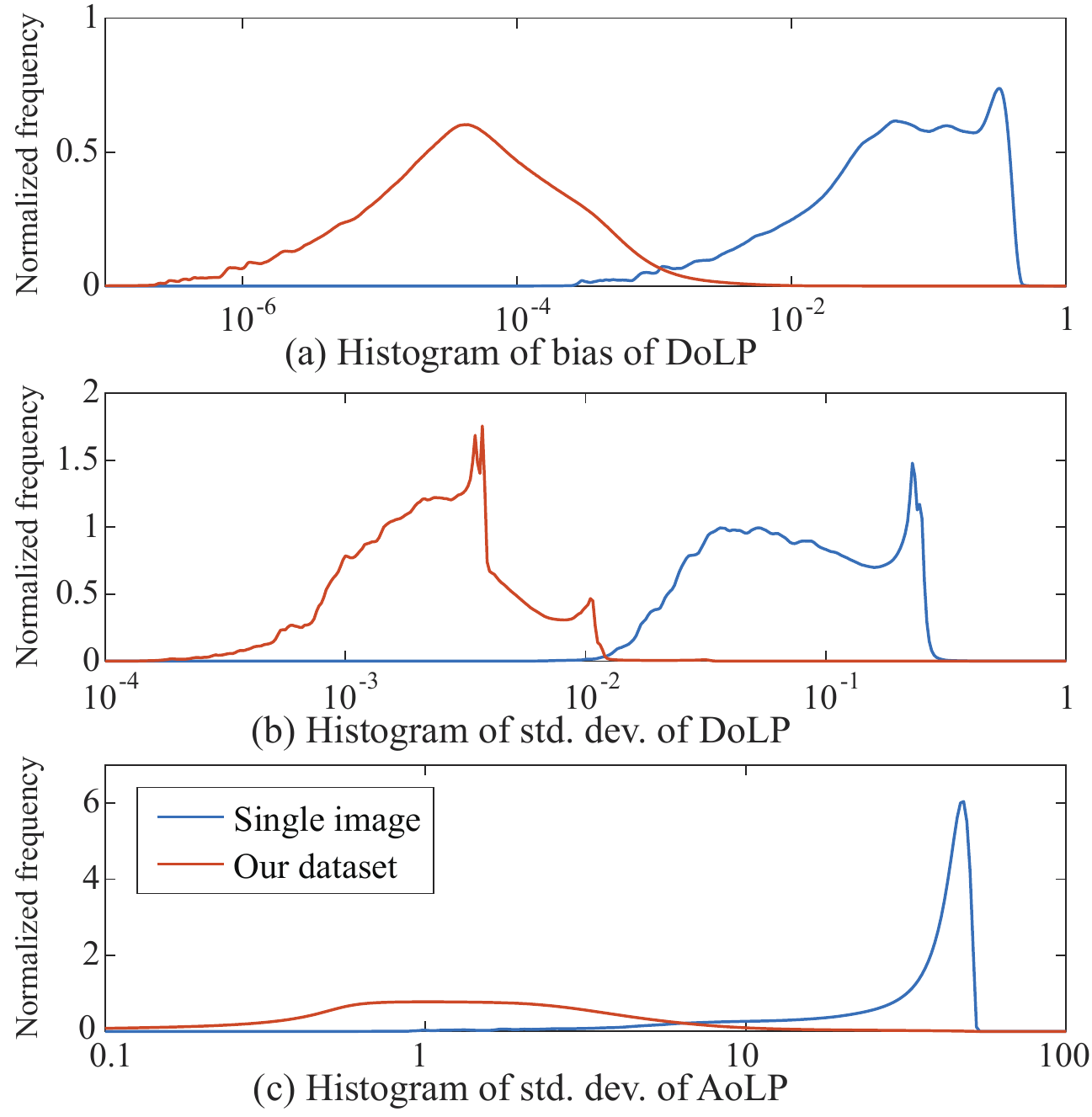}%
	\vspace{-1mm}%
	\caption{\label{fig:dataset_statistics_histogram}
	Histogram of the estimated DoLP bias, the DoLP standard deviation, and the AoLP standard deviation for the dataset. The histogram is computed in bins on a logarithmic scale, and the values are normalized to a probability density scale.}
	\vspace{-1mm}%
\end{figure}

Using our noise analysis model, we estimate the statistics of the physical values from our static scene dataset. The model treats the pixel values as Gaussian random variables in Equation~\eqref{eq:image_noise_model}. Additionally, we captured the static scene under static illumination, resulting in independent random variables with the same Gaussian distribution for pixel values over time at the same position. Through the arithmetic of Gaussian random variables, the variance of the mean from N independent Gaussian random variables reduces to 1/N. Similarly, we estimated each pixel's variance as the variance of a single pixel divided by the number of captured frames. From the mean values and estimated variances, DoLP and AoLP distributions can be reconstructed. Consequently, we accumulated the per-pixel statistics from the per-pixel DoLP and AoLP distributions into a histogram. The histogram and its summary are shown in Figure~\ref{fig:dataset_statistics_histogram} and Table~\ref{tab:dataset_statistics}. 

Although the PSNR of the $s_0$ from the single captures is reasonable for the image dataset, around 90\% of pixels exhibit over a 10-degree standard deviation in the AoLP, and more than 80\% of pixels display over a 0.01 bias in the DoLP. Our dataset minimizes noise through static burst captures, ensuring robustness in the polarization property domain, not just the intensity domain.

\begin{table}
	\centering
	\small
	\caption{\label{tab:dataset_statistics}%
	Dataset statistics predicted by the model indicate that the image-like component $s_0$ is represented as a general image metric, while the polarization characteristics are denoted as the percentage of pixels satisfying the model-based metric. Bold text emphasizes the result with the least noise in that context metric.}
	\vspace{-1mm}%
	\begin{tabular}{lllcc}
		\toprule
		& \multicolumn{2}{c}{Metric} & Single image & Our dataset \\
		\midrule
		$s_0$ & \multicolumn{2}{l}{PSNR} & 52.61 & \textbf{82.83}\\
		\midrule
		\multirow{4}{*}{DoLP} & Bias & $<$ 0.01 & 14.64\,\% & \textbf{\phantom{0}99.97}\,\% \\
		& & $<$ 0.001 & \phantom{0}1.47\,\% & \textbf{\phantom{0}97.85}\,\% \\[0.5pt]
		\cline{2-5} \\[-9pt]
		& Std.dev. & $<$ 0.1 & 64.20\,\% & \textbf{100.00}\,\% \\
		& & $<$ 0.01 & \phantom{0}0.10\,\% & \textbf{\phantom{0}97.30}\,\% \\
		\midrule
		\multirow{2}{*}{AoLP} & Std.dev. & $<$ 10\degree & 11.57\,\% & \textbf{\phantom{0}97.41}\,\% \\
		& & $<$ 5\degree & \phantom{0}4.83\,\% & \textbf{\phantom{0}91.86}\,\% \\
		\bottomrule
	\end{tabular}
		\vspace{-3mm}%
\end{table}

\section{Application}
\label{sec:application}

Our datasets make it possible to develop and validate learning-based burst super-resolution of polarization images. Therefore, we validate the neural network models in the polarization image domain. We select 5 state-of-the-art burst SR networks for comparison: MFIR \cite{Bhat_2021_ICCV}, BSRT \cite{luo2022bsrt}, Burstormer \cite{Dudhane_2023_CVPR}, FBANet \cite{Wei_2023_ICCV}, and BurstM \cite{Kang24}. The models are adopted for RGB images, so we modified the code as needed, such as altering the number of input and output channels, performing demosaicking if required, or using optical flow with $s_0$. Refer to the supplemental document for the training details. We validate which strategy performs better for burst SR in polarization images by comparing RGB pretrained models with those trained on polarization image datasets. Next, we compare burst SR networks using the PolarBurstSR dataset.

\begin{figure*}
\centering
\def\svgscale{1}
\footnotesize
\graphicspath{{figs/}}
\resizebox{1.0\linewidth}{!}{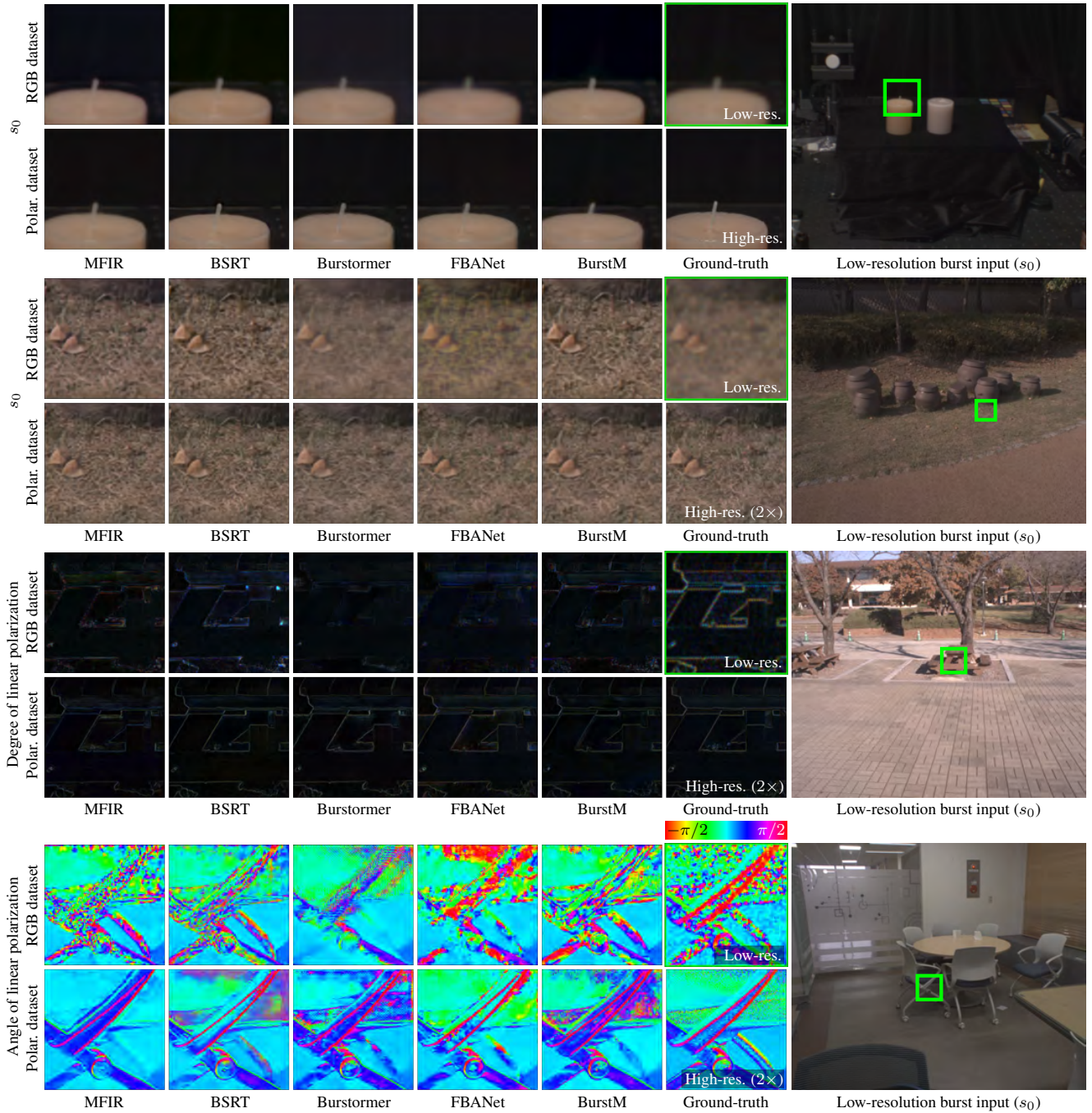}
\caption{Qualitative comparisons of 2$\times$ burst super-resolution results on the PolarNS dataset with various scenes conditions. Overall, the trained models in the polarization dataset in the lower row are sharper than those in the RGB dataset in the upper row and show results closer to GT. In particular, in AoLP, although much geometric information is lost due to noise in LR and RGB trained models, this information is reconstructed in the polarization-trained models.}
\label{fig:denoise_main}
\end{figure*}

\newcommand{\greentex}[1]{\textcolor{Green}{\footnotesize{#1}}}
\newcommand{\redtex}[1]{\textcolor{Red}{\footnotesize{#1}}}
\begin{table*}[t!]
\small
\begin{center}
\caption{\label{table:quantitative-results}
Quantitative evaluation. \textcolor{Green}{Green text} indicates improvement, while \textcolor{Red}{red text} denotes a performance decrease. We also mark the top score as \textbf{bold text}. The prefix \textit{p}- indicates that it is trained on a polarization dataset. In most metrics, models trained on polarization datasets perform significantly better.}
\vspace{-1mm}
\begin{tabular}{lccccccccccc}
	\toprule
	\multicolumn{1}{c}{} & \multicolumn{5}{c}{Synthetic polar dataset} & &\multicolumn{5}{c}{Real polar dataset} \\
	\midrule
	 \multicolumn{1}{c}{\multirow{2}{*}{Method}} & \multicolumn{3}{c}{$s_{0}$} & DoLP & AoLP & & \multicolumn{3}{c}{$s_{0}$} & DoLP & AoLP \\
	 & PSNR$\uparrow$ & SSIM$\uparrow$ & LPIPS$\downarrow$ & PSNR$\uparrow$ & PSNR$\uparrow$ & & PSNR$\uparrow$ & SSIM$\uparrow$ & LPIPS$\downarrow$ & PSNR$\uparrow$ & PSNR$\uparrow$ \\
	\midrule
	MFIR~\cite{Bhat_2021_ICCV} & 32.28 & 0.870 & 0.091 & 16.79 & 14.90 & & 43.16 & 0.962 & 0.084 & 23.76 & 16.67 \\
	\multirow{2}{*}{\textit{p}-MFIR} & 42.18 & 0.971 & 0.035 & 21.26 & 15.97 & & 45.18 & 0.977 & 0.063 & 22.00 & 15.74 \\[-2pt]
	& \greentex{+9.90}&\greentex{+0.101}&\greentex{-0.056}&\greentex{+4.47}&\greentex{+1.07}& &\greentex{+2.02}&\greentex{+0.015}&\greentex{-0.021}&\redtex{-1.76}&\redtex{-0.93} \\[-2pt]
	\midrule
	BSRT~\cite{luo2022bsrt} & 31.95 & 0.860 & 0.067 & 20.70 & 14.29 & & 44.15 & 0.973 & 0.073 & 23.03 & 16.14  \\ 
	\multirow{2}{*}{\textit{p}-BSRT} & \textbf{43.75} & \textbf{0.979} & \textbf{0.021} & 20.31 & \textbf{17.09} & & \textbf{45.69} & \textbf{0.979} & \textbf{0.051} & 25.17 & 17.06 \\ [-2pt]
	& \greentex{+11.80}& \greentex{+0.119}& \greentex{-0.046}& \redtex{-0.39}& \greentex{+2.80}& & \greentex{+1.54}& \greentex{+0.006}& \greentex{-0.022}& \greentex{+2.14}& \greentex{+0.92} \\[-2pt]
	\midrule
	Burstormer~\cite{Dudhane_2023_CVPR} & 32.03 & 0.862 & 0.071 & 21.36 & 15.16 & & 40.72 & 0.938 & 0.147 & 24.90 & 16.95  \\ 
	\multirow{2}{*}{\textit{p}-Burstormer} & 41.12 & 0.965 & 0.049 & \textbf{24.49} & 16.79 & & 43.29 & 0.973 & 0.061 & 25.84 & 17.36 \\ [-2pt]
	& \greentex{+9.09}& \greentex{+0.103}& \greentex{-0.022}& \greentex{+3.13}& \greentex{+1.63}& &\greentex{+2.57}& \greentex{+0.035}& \greentex{-0.086}& \greentex{+0.94}& \greentex{+0.41}
	 \\[-2pt]
	\midrule
	FBAnet~\cite{Wei_2023_ICCV} & 33.94 & 0.892 & 0.201 & 15.29 & 13.20 & & 40.57 & 0.946 & 0.138 & 24.28 & 16.14 \\
	\multirow{2}{*}{\textit{p}-FBAnet} & 36.08 & 0.914 & 0.193 & 21.23 & 16.88 & & 45.64 & 0.976 & 0.055 & \textbf{26.92} & \textbf{18.14} \\[-2pt]
	&\greentex{+2.14}&\greentex{+0.018}&\greentex{-0.008}&\greentex{+5.94}&\greentex{+3.68}&&\greentex{+5.07}&\greentex{+0.030}&\greentex{-0.083}&\greentex{+2.64}&\greentex{+2.00}
	 \\  [-2pt]
	\midrule
	BurstM~\cite{Kang24} & 32.00 & 0.860 & 0.067 & 21.77 & 15.37 & & 43.82 & 0.972 & 0.075 & 22.78 & 16.64 \\ 
	\multirow{2}{*}{\textit{p}-BurstM} & 39.42 & 0.953 & 0.063 & 21.34 & 16.08 & & 45.19 & 0.975 & 0.063 & 23.67 & 16.32 \\[-2pt]
	& \greentex{+7.42}&\greentex{+0.093}&\greentex{-0.004}&\redtex{-0.43}&\greentex{+0.71}&&\greentex{+1.37}&\greentex{+0.003}&\greentex{-0.012}&\greentex{+0.89}&\redtex{-0.32}
	 \\[-2pt]
	\bottomrule
\end{tabular}
\end{center}
\vspace{-5mm}
\end{table*}

\subsection{Comparison with RGB pretrained model}
As a validation of the training in the polarization image domain, we compare pre-trained models on RGB images with trained models on the polarization image dataset. The input channels, output channels, and Bayer patterns differ between RGB and polarization images; therefore, the RGB pre-trained model cannot be applied directly. We considered the differences in the Bayer patterns to adapt the RGB pre-trained model for use with polarization raw images. The polarization Bayer pattern for each of the 0\degree, 45\degree, 90\degree, and 135\degree linear polarization angles can be represented as a CFA Bayer pattern with two pixels dilation. Each angle's linear polarization image follows the super-resolution process like an RGB raw image, and they are aligned by considering their original positions in the polarization Bayer pattern.

The results are presented in Figure~\ref{fig:denoise_main} and Table~\ref{table:quantitative-results}. The models trained with polarization are designated by the prefix p- in the table. Training on the polarization dataset demonstrates performance improvements across most models and metrics, regardless of whether the dataset is real or synthetic. It produces clearer images for the $s_0$ components and more stable images in the DoLP images, while RGB pre-trained models yield varying shapes depending on the specific model. 
The comparison in the AoLP image shows a more significant difference. The low-resolution input fails to generate a normal signal due to noise interference, and the RGB pre-trained model shows some geometric features but lacks sufficient information to reconstruct the geometry. Conversely, the polarization model, like GT, effectively reflects the scene's geometric information. Overall, the results highlight that the contribution of information propagation among each polarization image is more significant than the size of the training dataset.

\subsection{Comparison among burst SR models}
We compare the burst SR models in the PolarBurstSR dataset as a benchmark. The results are presented in Figure~\ref{fig:denoise_main} and Table~\ref{table:quantitative-results}. BSRT achieves the best result for the metric $s_0$, which closely resembles the RGB image. However, it does not perform best regarding the polarization characteristics, DoLP and AoLP. FBANet excels in the polarization characteristics, ranking second or third in $s_0$. Nevertheless, in the qualitative comparison, the advantage in the polarization characteristics is not clearly demonstrated, while the $s_0$ results show BSRT's clarity. This may be due to the absence of a metric to assess the significance of DoLP or AoLP results, as PSNR, a metric reliant on mean squared error, is employed.

\section{Conclusion}
\label{sec:conclusion}
We introduce PolarNS and PolarBurstSR, two novel datasets designed to enhance noise analysis and burst super-resolution in polarization imaging. PolarNS offers a detailed characterization of polarization noise statistics, enabling precise modeling and evaluation of noise propagation. Our polarization noise analysis model further quantifies noise behavior, providing a rigorous framework for assessing the confidence of polarization-derived properties.
As an application, we demonstrate the effectiveness of burst super-resolution for polarization imaging by comparing models trained on conventional RGB datasets with those trained on our dedicated polarization dataset. The results underscore the importance of polarization-specific training in improving both intensity image resolution and the accuracy of polarization properties.
By publicly releasing our datasets, pretrained models, and training pipelines, we aim to establish a benchmark for polarization burst super-resolution and promote future research in the denoising and high-resolution reconstruction of polarization images.

\clearpage
{
    \small
    \bibliographystyle{ieeenat_fullname}
    \bibliography{bibliography,bibliography_sig22}
}


	\onecolumn
	{
		\centering
		\Large
		\textbf{\thetitle}\\
		\vspace{0.5em}Supplementary Material \\
		\vspace{1.0em}
	}
	\section{Polarization Super-resolution Image Processing Details}
The camera used is the same model as the one used to obtain the noise statistics. To achieve a 4$\times$ super-resolution configuration like that of the BurstSR dataset, we capture the burst images with an 8mm lens, which has a focal length approximately four times shorter. The reference frame is set to the first frame of the burst image. The ground truth image is debayered into 12 channel images, consisting of three colors multiplied by four polarization angles, using bilinear interpolation. The polarization angle images are treated as occupying the same pixel locations, which results in the spatial resolution of the ground truth being halved at this stage. The homography between the ground truth and the reference frame is calculated using the SIFT feature~\cite{Lowe2004} and RANSAC matching~\cite{Fischler1981}. The ground truth images are warped to align with the reference frame of the burst images using the calculated homography. While warping the image pixels, the $s_1$ and $s_2$ values are adjusted based on the rotation from the homography. Due to the multiple device acquisitions, the burst frames and ground truth images exhibit misalignments caused by disparity and color differences. We exclude image pairs with a normalized cross-correlation lower than 0.9 and adjust the image colors using a per-image color correction matrix. The images are cropped to sizes of 192$\times$192 and 384$\times$384 for frames, and 384$\times$384 for the ground truth, resulting in an overall super-resolution of 2$\times$. The actual dimensions for input and output in the networks are 48$\times$48$\times$16 and 384$\times$384$\times$9 respectively, yielding a spatial resolution ratio of 8$\times$, which aligns with the original BurstSR and many burst super-resolution models.

\section{Training Details}

Training the network requires a massive labeled dataset. However, acquiring a real burst image dataset involves high costs and significant effort, making it challenging to collect a sufficient amount of data through real captures. Instead, we generate synthetic data from other unlabeled polarization image datasets. We first train the network using synthetic data and then fine-tune it with real image data. 

\mparagraph{Synthetic data training}
In the synthetic data training, we follow the synthetic burst image generation method outlined in DBSR\cite{Bhat2021}. We generate synthetic burst polarization images from the RSP dataset\cite{Kurita_2023_WACV}. The RSP dataset includes 1586 and 176 synthetic polarization images in the training and validation sets, respectively, along with 238 carefully captured real polarization images in the test set. We maintain the original order and use their separation directly. Unlike DBSR, the RSP dataset is already in linear RGB color space, so we do not apply the unprocessing pipeline\cite{brooks2019unprocessing} to extract raw pixel values. For generation, we first randomly crop the image to obtain the labeled ground truth and the 0-th frame. For the other frames, we include additional random translations and rotations within the ranges of [-24, 24] pixels and [-1, 1] degree. Since the input to FBANet\cite{Wei_2023_ICCV} consists of aligned images, we do not introduce random movement for FBANet in the synthetic data training. Subsequently, the frames are downsampled by a factor of 2 and masked using a polarization Bayer pattern that matches the off-the-shelf polarization image sensor. Finally, random synthetic noise is added to each pixel based on the sensor noise distribution defined by Brooks et al.\cite{brooks2019unprocessing}. The final ground truth has 384$\times$384$\times$12 that means rows$\times$columns$\times$RGB and polarization channels, while each frame has 48$\times$48$\times$16 that means rows of Bayer pattern lattices $\times$columns of that$\times$variations of the polarization Bayer pattern.

For actual training, we use 14 generated burst frames for each labeled sample. We implement the network in PyTorch. The network is trained using the ADAM optimizer with a fixed learning rate of 1e-4 for 500,000 iterations, with a batch size of 12. The loss function used is the L1 loss, which is commonly employed in previous burst SR works.

\mparagraph{Real data training}
The frames and GTs in the real burst SR dataset have positional misalignment and color differences resulting from being captured by two different devices in varying positions \cite{Wei_2023_ICCV}. Our dataset also has these intrinsic problems, so we use the aligned losses and metrics proposed in DBSR \cite{Bhat2021}. The original metric is developed for RGB images; therefore, the calculation of the optical flow, color mapping, and validation mask utilizes $s_0$ images. Training begins with the trained network from the synthetic data, and the network is fine-tuned over 50k iterations. The other settings remain the same as those used in synthetic data training.

\section{Validation of Stokes vector noise model}
\begin{figure*}[htpb]
	\centering
	\def\svgscale{1}
	\graphicspath{{figs/}}
	\resizebox{1.0\linewidth}{!}{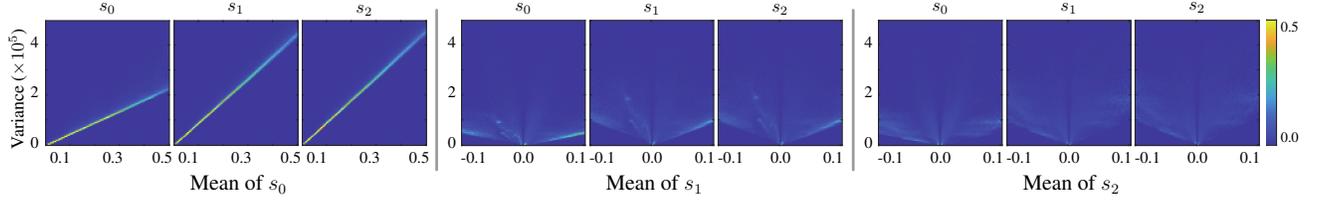}
	\caption{Histogram of noise for each component based on the values of the Stokes vector components. The noise of $s_0$, $s_1$, and $s_2$ is strongly correlated with the $s_0$ value.}
	\label{fig:stokes_histogram}
\end{figure*}

To demonstrate the dependency of the noise in the Stokes vector components and the true $s_0$ value, we obtain a histogram of the signal-to-noise ratio for each Stokes vector component. We calculate the variance of the observed values to illustrate the linear relationship with the true value. Histograms are generated for the three Stokes vector components of signal and noise, resulting in a total of nine histograms. The results are presented in Figure \ref{fig:stokes_histogram}. The rows represent the variance of each component, indicating the noise, while the columns denote the mean of each component, reflecting the signal. 
The first column clearly demonstrates the linear relationship between the true $s_0$ value and each Stokes vector component.
In contrast, the histograms comparing the noise with the $s_1$ and $s_2$ signals do not show a clear dependency in the second and third columns. In contrast, the histograms comparing the noise with the $s_1$ or $s_2$ signals do not exhibit a clear dependency in the second and third columns.

	\section{Detail derivation process of the AoLP distribution}
The noise model of AoLP is the marginal distribution of the phase of ${\hat{s}}_1$ and ${\hat{s}}_2$. It can be represented by the integration of polar coordinate, which is expressed as:
\begin{align}
	f_{\phi }^{{}}\left( 2\hat{\phi }|{{s}_{1}},{{s}_{2}},\sigma _{v}^{2} \right)&=\int_{0}^{\infty }{f\left( {{{\hat{s}}}_{1}},{{{\hat{s}}}_{2}} \right)rdr} \nonumber \\ 
	& =\int_{0}^{\infty }{f\left( r\cos 2\hat{\phi },r\sin 2\hat{\phi } \right)rdr} \nonumber
\end{align}
We expand the Gaussian distribution and organize it using the angle addition and subtraction of trigonometric function, it is expressed as follows:
\begin{align}
	f_{\phi }^{{}}\left( 2\hat{\phi }|{{s}_{1}},{{s}_{2}},\sigma _{v}^{2} \right)&=\int_{0}^{\infty }{\frac{1}{2\pi \sigma _{v}^{2}}\exp \left( -\frac{{{\left( r\cos 2\hat{\theta }-{{s}_{1}} \right)}^{2}}+{{\left( r\sin 2\hat{\theta }-{{s}_{2}} \right)}^{2}}}{2\sigma _{v}^{2}} \right)rdr}\nonumber \\ 
	&=\int_{0}^{\infty }{\frac{1}{2\pi \sigma _{v}^{2}}\exp \left( -\frac{{{\left( r\cos 2\hat{\phi }-{{s}_{pol}}\cos 2\phi  \right)}^{2}}+{{\left( r\sin 2\hat{\phi }-{{s}_{pol}}\sin 2\phi  \right)}^{2}}}{2\sigma _{v}^{2}} \right)rdr} \nonumber \\
	&=\int_{0}^{\infty }{\frac{1}{2\pi \sigma _{v}^{2}}\exp \left( -\frac{{{r}^{2}}+{{s}_{pol}}^{2}-2r{{s}_{pol}}\left( \cos 2\hat{\phi }\cos 2\phi +\sin 2\hat{\phi }\sin 2\phi  \right)}{2\sigma _{v}^{2}} \right)rdr} \nonumber\\
	&=\int_{0}^{\infty }{\frac{1}{2\pi \sigma _{v}^{2}}\exp \left( -\frac{{{r}^{2}}+{{s}_{pol}}^{2}-2r{{s}_{pol}}\left( \cos \left( 2\hat{\phi }-2\phi  \right) \right)}{2\sigma _{v}^{2}} \right)rdr} \nonumber
\end{align}
From the integral above, we can see that the ratio of the intensity of the polarization component and Stokes vector noise,${s}_{pol}/\sigma _{v}$, is a parameter of the function.
Using the following substitution $t=\frac{r}{\sigma _{v}^{{}}}$, $dt=\frac{1}{\sigma _{v}^{{}}}dr$, and $A=\frac{1}{2\pi }\exp \left( -\frac{{{s}_{pol}}_{{}}^{2}{{\sin }^{2}}\left( 2\hat{\phi }-2\phi  \right)}{2\sigma _{v}^{2}} \right)$, the equation is summarized as follows:
\begin{align}
	f_{\phi }^{{}}\left( 2\hat{\phi }|{{s}_{1}},{{s}_{2}},\sigma _{v}^{2} \right)
	&=\int_{0}^{\infty }{\frac{1}{2\pi }\exp \left( -\frac{{{t}^{2}}}{2}-\frac{{{s}_{pol}}^{2}}{2\sigma _{v}^{2}}-\frac{{{s}_{pol}}t\left( \cos \left( 2\hat{\phi }-2\phi  \right) \right)}{\sigma _{v}^{{}}} \right)tdt}\nonumber \\ 
	&=\int_{0}^{\infty }{\frac{1}{2\pi }\exp \left( -\frac{1}{2}{{\left( t-\frac{{{s}_{pol}}\cos \left( 2\hat{\phi }-2\phi  \right)}{\sigma _{v}^{{}}} \right)}^{2}}-\frac{{{s}_{pol}}_{{}}^{2}{{\sin }^{2}}\left( 2\hat{\phi }-2\phi  \right)}{2\sigma _{v}^{2}} \right)tdt} \nonumber \\ 
	&=\int_{0}^{\infty }{At\exp \left( -\frac{1}{2}{{\left( t-\frac{{{s}_{pol}}\cos \left( 2\hat{\phi }-2\phi  \right)}{\sigma _{v}^{{}}} \right)}^{2}} \right)dt} \nonumber 
\end{align}
Finally, by performing integration using the differentiation $\frac{d}{dx}\exp \left( -\frac{1}{2}{{x}^{2}} \right)=-x\exp \left( -\frac{1}{2}{{x}^{2}} \right)$ and solving the substitution, it is expressed as follows:
\begin{align}
	f_{\phi }^{{}}\left( 2\hat{\phi }|{{s}_{1}},{{s}_{2}},\sigma _{v}^{2} \right)
	& =\int_{0}^{\infty }{A\left( t-\frac{{{s}_{pol}}\cos \left( 2\hat{\phi }-2\phi  \right)}{\sigma _{v}^{{}}} \right)\exp \left( -\frac{1}{2}{{\left( t-\frac{{{s}_{pol}}\cos \left( 2\hat{\phi }-2\phi  \right)}{\sigma _{v}^{{}}} \right)}^{2}} \right)dt} \nonumber \\ 
	& +\int_{0}^{\infty }{\frac{A{{s}_{pol}}\cos \left( 2\hat{\phi }-2\phi  \right)}{\sigma _{v}^{{}}}\exp \left( -\frac{1}{2}{{\left( t-\frac{{{s}_{pol}}\cos \left( 2\hat{\phi }-2\phi  \right)}{\sigma _{v}^{{}}} \right)}^{2}} \right)dt} \nonumber \\
	&=A\exp \left( -\frac{{{s}_{pol}}_{{}}^{2}\cos _{{}}^{2}\left( 2\hat{\phi }-2\phi  \right)}{2\sigma _{v}^{2}} \right)+\frac{\sqrt{2\pi }A{{s}_{pol}}\cos \left( 2\hat{\phi }-2\phi  \right)}{\sigma _{v}^{{}}}\Phi \left( \frac{{{s}_{pol}}\cos \left( 2\hat{\phi }-2\phi  \right)}{\sigma _{v}^{{}}} \right) \nonumber \\
	&=\frac{1}{2\pi }\exp \left( -\frac{{{s}_{pol}}_{{}}^{2}}{2\sigma _{v}^{2}} \right)+\frac{\psi \cos \left( 2\hat{\phi }-2\phi  \right)}{\sqrt{2\pi }\sigma _{v}^{{}}}\exp \left( -\frac{{{s}_{pol}}_{{}}^{2}{{\sin }^{2}}\left( 2\hat{\phi }-2\phi  \right)}{2\sigma _{v}^{2}} \right)\Phi \left( \frac{{{s}_{pol}}\cos \left( 2\hat{\phi }-2\phi  \right)}{\sigma _{v}^{{}}} \right) \nonumber
\end{align}
where $\Phi$ is the cumulative distribution function of the standard normal distribution.

	\section{Discussion}
\label{sec:discussion}

Although we build the dataset and validate the noise model as elaborately as possible, there are still several considerable tasks left for future work.

\paragraph{Noise comparison among dataset}
A single capture of the polarization image may not provide sufficient effective polarization information unless its noise level is sufficiently low. Therefore, we proposed to present the noise statistics data for the first time. As a result, we couldn't compare our noise model and physical values in polarization with the previous dataset; instead, we only compared using noise level estimation from a single image in the $s_0$ domain. However, this comparison is not an ideal way to assess how robust the polarization data is against noise. 

The model for single image noise detection is not the shot and read noise model; it is the additive white Gaussian noise model. This model creates a gap between the effectiveness of the data's polarization properties and the quality of the estimated noise level. The estimated noise level increases as the overall intensity of the images becomes larger because shot noise depends on the number of photons arriving at the sensor. However, the SNR of $s_0$ or $s_pol$, that are key parameters for the effectiveness of polarimetric properties, increases when the overall captured intensity is larger. Therefore, this metric might not distinguish between sophisticated captures of bright scenes and noisy scenes with underexposure. For example, the KAUST polarization image dataset has the highest estimated noise level, suggesting that the capture setup of the KAUST dataset could be the best among the comparison datasets. They captured images using a DSLR with a rotating polarizer, 2$ imes$2 pixel binning, and an integration of 100 burst images.

We provide noise statistics, equipment specifications, and parameter settings. Additionally, new comparison methods for evaluating the effectiveness of polarimetric data without provided noise statistics could also be considered for future work.

\paragraph{Limitation of the noise model}
We adopted a shot-and-read noise model with a Gaussian distribution as the base noise model in the sensor domain for deriving the noise distribution of the polarization properties. The assumption of Gaussian noise is very useful for derivation, and it yields quite good results in many cases. However, it can deviate from the model when the image has low intensity.
First, the effect of quantization noise increases when the digital number becomes smaller. 
Second, the distribution of read noise does not strictly follow a Gaussian distribution.
For instance, in Figure 4 of the main paper, the high SNR aligns well with the model, but the low SNR displays a spiky shape, despite the entire distribution adhering to our model.
This might be the effect of quantization noise, which can limit the number of cases of the DOLP and AOLP in low-intensity data. A more sophisticated model for low intensity could be future work.

\paragraph{Scene variety}
We reduce noise using massive images in the temporal domain. This approach effectively and predictably minimizes noise.
However, it limits scene variety due to the long acquisition time.
For outdoor and indoor scene capture, we capture around 1,000 images over approximately 2 minutes. During this capture time, the scene must remain motionless and free of illumination changes. Therefore, our dataset excludes objects affected by their own movement or wind, such as plants, animals, clouds, rivers, or lakes. Moreover, the illumination conditions must also be static, so we cannot capture images in rainy, snowy, cloudy, or foggy weather, nor during sunrise or sunset, even though these varying weather conditions may produce meaningful polarization effects from scattering or large incident angles. The acquisition method for a polarization noise-reduced image under various objects and illumination conditions remains future work.

\paragraph{Incoherency between burst images and GT}
As criticized by Wei et al.~\cite{Wei_2023_ICCV}, burst SR datasets acquired with multiple devices simultaneously have incoherencies between different devices, as does the PolarBurstSR dataset. Burst SR datasets include two types of images in a sample: hand-held burst images and a GT image. These should have different focal lengths, so datasets select temporal multiplexing~\cite{Wei_2023_ICCV,cai2019toward} or multiple device capturing. Acquiring on multiple devices inherently causes misalignment from different views and color differences among devices~\cite{Bhat2021}. Despite these problems, we captured images using two cameras with different lenses. Due to the susceptibility of the polarization properties to noise, we captured around 1000 images for GT using a tripod. Additionally, since the polarization cameras did not perform like commercial DSLR cameras, we cannot use zoom lenses for different focal lengths in a single optical system. Therefore, temporal multiplexing requires the mounting and unmounting of elements in the setup, which could undermine the benefits of single device acquisition.

	\clearpage

\section{Synthetic dataset results}
\begin{figure*}[htpb]
	\centering
	\def\svgscale{1}
	\footnotesize
	\graphicspath{{figs/}}
	\resizebox{1.0\linewidth}{!}{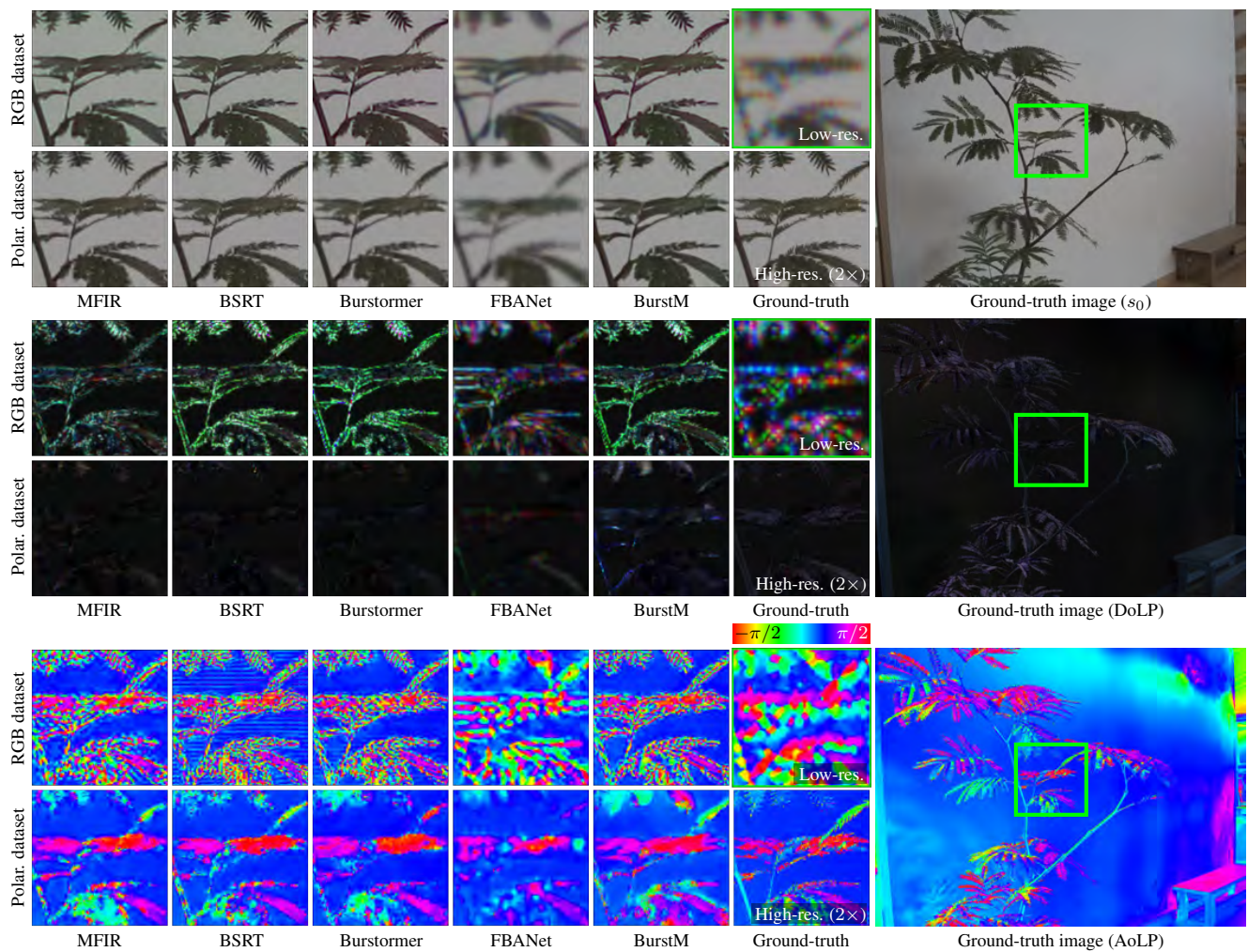}
	\caption{{Qualitative comparisons on the polarization synthetic dataset.}}
	\label{fig:denoise_sup0}
\end{figure*}
\clearpage
\begin{figure*}[htpb]
	\centering
	\def\svgscale{1}
	\footnotesize
	\graphicspath{{figs/}}
	\resizebox{1.0\linewidth}{!}{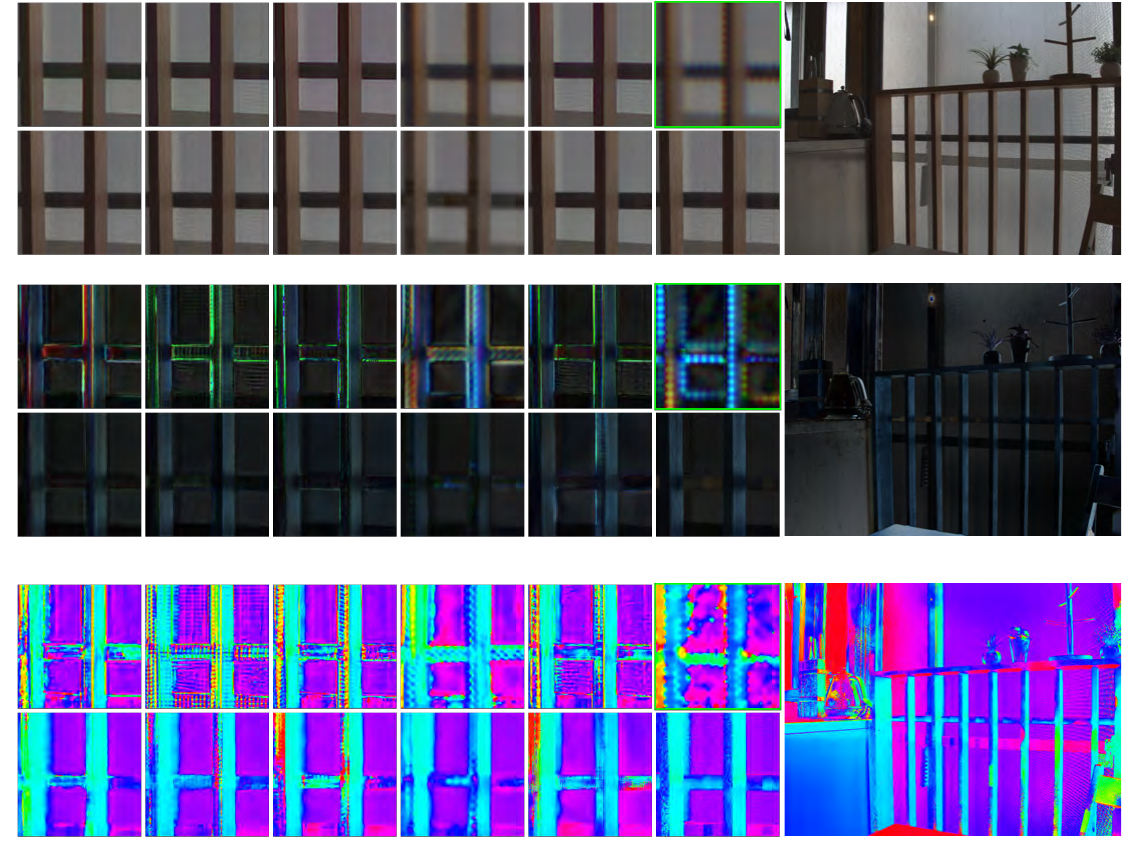}
	\caption{{Qualitative comparisons on the polarization synthetic dataset.}}
	\label{fig:denoise_sup1}
\end{figure*}
\clearpage
\begin{figure*}[htpb]
	\centering
	\def\svgscale{1}
	\footnotesize
	\graphicspath{{figs/}}
	\resizebox{1.0\linewidth}{!}{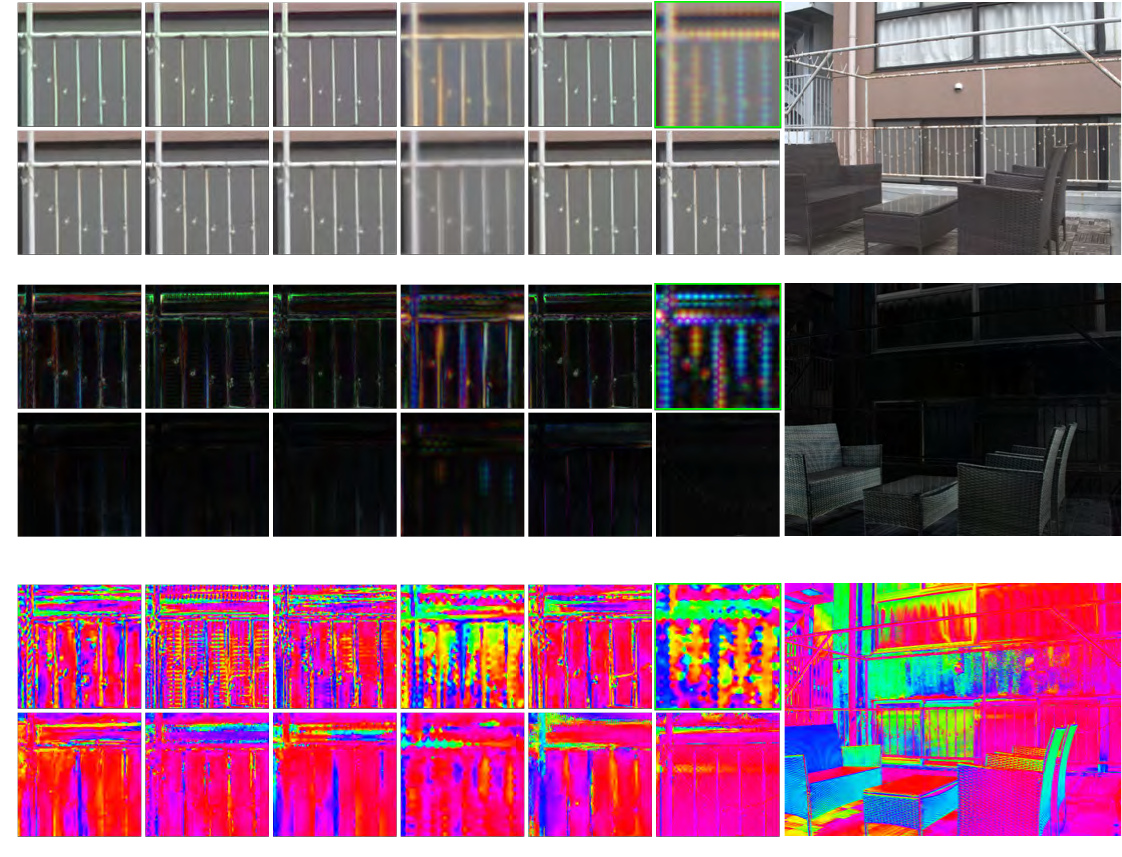}
	\caption{{Qualitative comparisons on the polarization synthetic dataset.}}
	\label{fig:denoise_sup2}
\end{figure*}
\clearpage
\begin{figure*}[htpb]
	\centering
	\def\svgscale{1}
	\footnotesize
	\graphicspath{{figs/}}
	\resizebox{1.0\linewidth}{!}{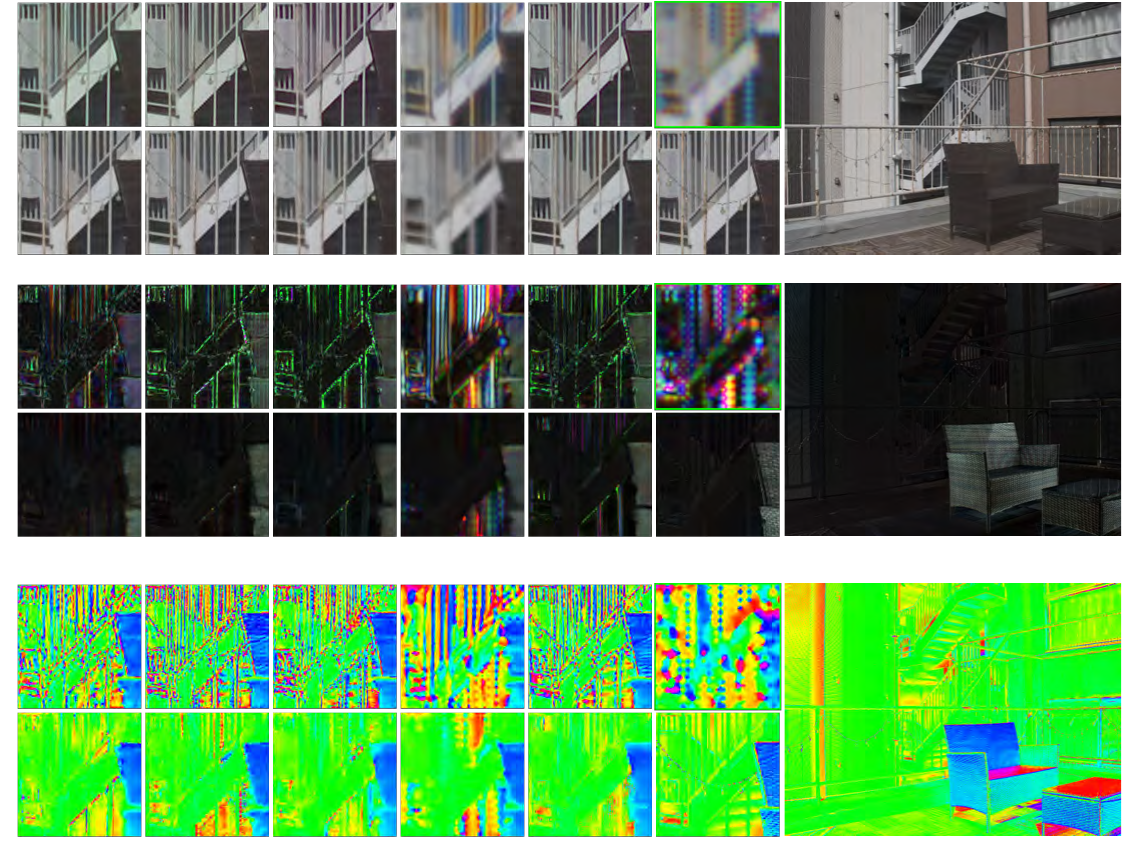}
	\caption{{Qualitative comparisons on the polarization synthetic dataset.}}
	\label{fig:denoise_sup3}
\end{figure*}
\clearpage

	{
		\small
		\bibliographystyle{ieeenat_fullname}
		\bibliography{bibliography}
	}

\end{document}